\documentclass[twoside]{wujart}

\usepackage[T1]{fontenc}
\usepackage[utf8]{inputenc}

\usepackage{hyperref}
\usepackage{amsmath,amsfonts,amssymb,amsthm,dsfont}
\usepackage{booktabs}

\usepackage{graphicx}
\usepackage{subfigure} 
\usepackage{enumerate}

\theoremstyle{remark}

\newcommand\il[1]{\big\langle #1 \big\rangle}

\def\Z{\mathcal{Z}}

\def\R{\mathbb{R}}

\def\D{\mathcal{D}}
\def\E{\mathcal{E}}

\def\1F1{\mbox{$_{1}{F}_{\!1}$}}

\usepackage{color}

\begin{document}

\author{Szymon Knop Marcin Mazur  Jacek Tabor \\Igor Podolak Przemys\l{}aw Spurek \\{\small\rm Faculty of Mathematics and Computer Science \\ Jagiellonian University, \L{}ojasiewicza 6, 30-348 Krak\'ow, Poland \\
e-mail: \textit{szymon.knop@doctoral.uj.edu.pl } }}


\title{{\large\bf Sliced generative models}}

\maketitle

\abstract{In this paper we 
discuss a class of  AutoEncoder based generative models based on one dimensional sliced approach. The idea is based on the reduction of the discrimination between samples to one-dimensional case.
Our experiments show that methods can be divided into two groups. First consists of methods which are a modification of standard normality tests, while the second is based on classical distances between samples. 
It turns out that both groups are correct generative models, but the second one gives a slightly faster decrease rate of Fr\'{e}chet Inception Distance (FID).
}


\section{Introduction}


In recent years a number of generative models based on  AutoEncoder architecture were constructed (see, e.g.,~\cite{kingma2014auto,kolouri2018sliced,tabor2018cramer,tolstikhin2017wasserstein}). Some of them have applied elegant geometric properties of the optimal transport (OT) problem and the Wasserstein distances. An important example is given in~\cite{kolouri2018sliced}, where the authors construct Sliced-Wasserstein  AutoEncoder  (SWAE) -- a generative model that performs well without the need for training an adversarial network but, on the other hand, with necessity of sampling from the prior distribution $P_\Z$ on the latent $\Z$. 
Specifically, the method applied there uses the sliced Wasserstein distance between the distribution of encoded training samples $(z_i)$ and $P_\Z$~ \cite{kolouri2018sliced}. SWAE has an efficient numerical solution that provides similar capabilities to Wasserstein  AutoEncoders (WAE-MMD)~\cite{tolstikhin2017wasserstein} and Variational  AutoEncoders~\cite{kingma2014auto}. A typical choice for $P_Z$ is the Gaussian distribution $N(0,1)$ even though SWAE is valid for any prior distribution. In this case, there is no need to sample from $P_\Z$, as long as we can analytically calculate a closed formula for the distance between a given sample $(x_i)$ and $N(0,1)$.

In our paper we follow the idea of~\cite{kolouri2018sliced} and make a comparison of few  AutoEncoder based generative models, for which the loss functions are given by appropriately chosen sliced distanced between $(z_i)$ and $N(0,1)$ that can be expressed in a closed form. Specifically, we use respective one-dimensional ``measures of normality'', including the 2nd Wasserstein~\cite{kolouri2018sliced} or the Cramer-Wold~\cite{tabor2018cramer} distances, as well those derived from some classical one dimensional goodness of fit tests for normality, i.e the Cram\'er-von Mises and the Kolmogorov-Smirnov. Let us also note that our approach is, up to some extent, related to that of~\cite{palmer2018reforming}, where the authors propose a method for training generative  AutoEncoders by explicitly testing $P_\Z$ via the Shapiro-Wilk test for (one-dimensional) normality, applied to a ``vectorized'' (multidimensional) sample~$(z_i)$.

Consequently, we use the following models:
\begin{enumerate}[-]\itemsep0.1pt
\item Sliced Wasserstein  AutoEncoder (SWAE)~\cite{kolouri2018sliced},
\item Sliced Closed Form Wasserstein  AutoEncoder (SCFWAE) -- an upgrade of SWAE,
\item Sliced Cramer-Wold  AutoEncoder (SCWAE), based on one dimensional Cramer-Wold distance~\cite{tabor2018cramer},
\item Sliced Cram\'er-von Mises  AutoEncoder (SCvMAE)using Cram\'er-von Mises normality test,
\item Sliced Kolmogorov-Smirnov  AutoEncoder (SKSAE), based on Kolmogorov-Smirnov normality test.
\end{enumerate}

There is also an important novelty which we have adopted from~\cite{tabor2018cramer}, namely we use the logarithm-like modification of the cost function. The main idea is that instead of considering the cost function of the form
$$
\mathrm{RecError}+\lambda \cdot \mathrm{NormalityIndex},
$$
which needs a grid search over $\lambda$ for the proper weighting of reconstruction error $\mathrm{RecError}$ and divergence from normality we can, typically with similar or better results, use
$$
\mathrm{RecError}+\log (\mathrm{NormalityIndex}).
$$
Thanks to this formulation, the cost function, from the optimization point of view, does not change with rescaling of the normality index by a constant $\lambda$ (in this case cost functions differ only by a constant $\log \lambda$, which results in the same gradient).

\begin{figure}[htb]
\centering
\begin{tabular}{@{}c@{}c@{}c@{}c@{}}
\rotatebox{90}{ \qquad\qquad SCvMAE} & \,
\includegraphics[height=3cm]{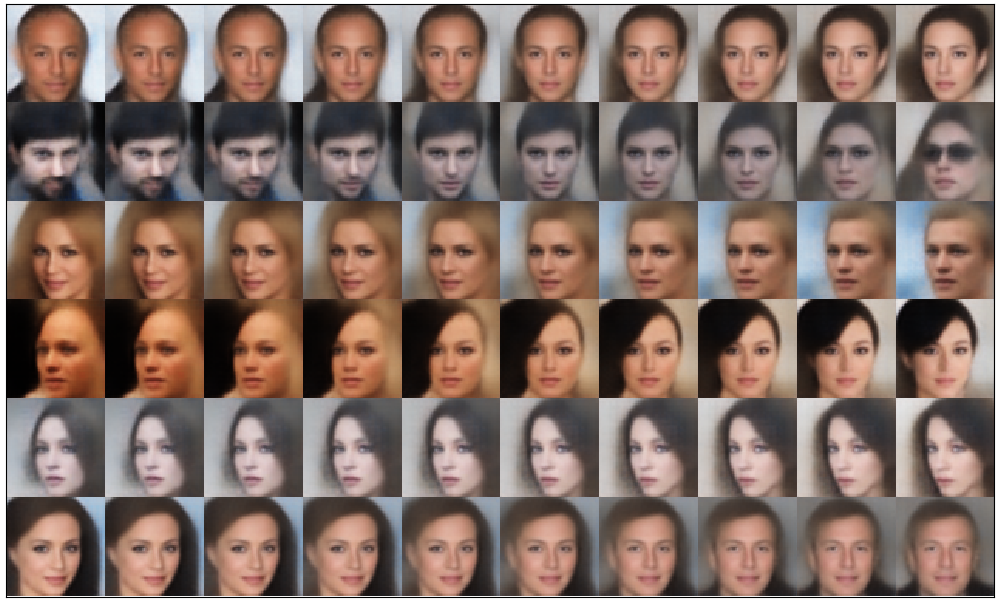} & \,
\includegraphics[height=3cm]{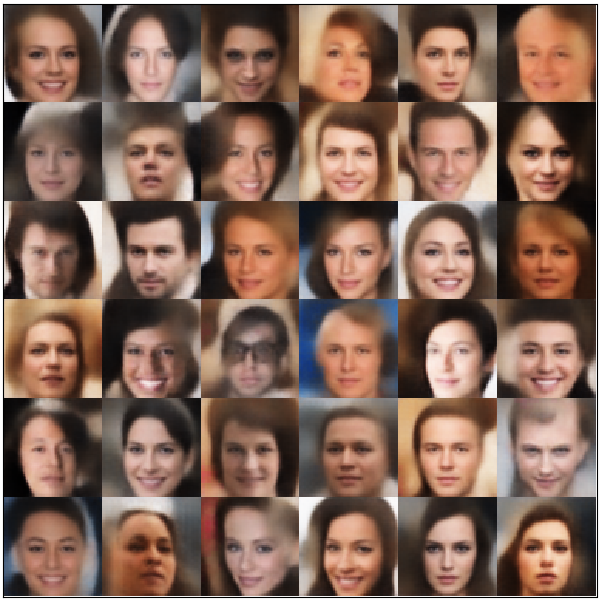}& \, 
\includegraphics[height=3cm]{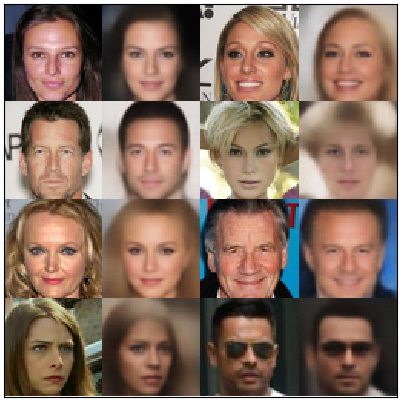} \\  
\rotatebox{90}{ \qquad\qquad SKSAE} & \,
\includegraphics[height=3cm]{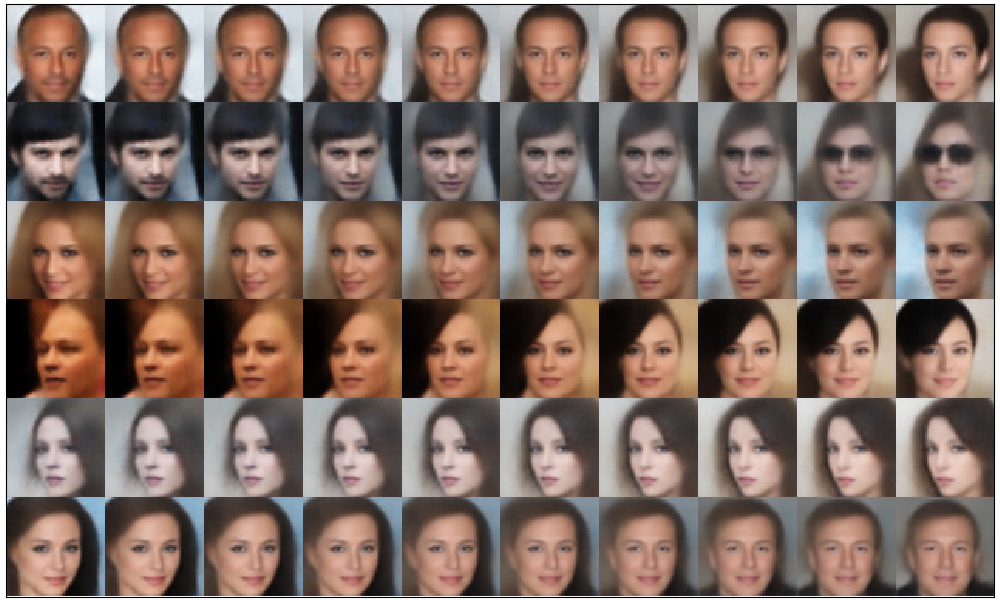} & \,
\includegraphics[height=3cm]{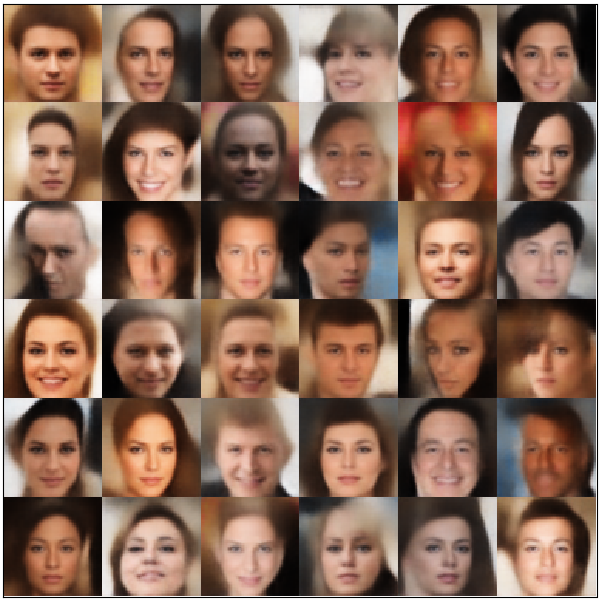}& \, 
\includegraphics[height=3cm]{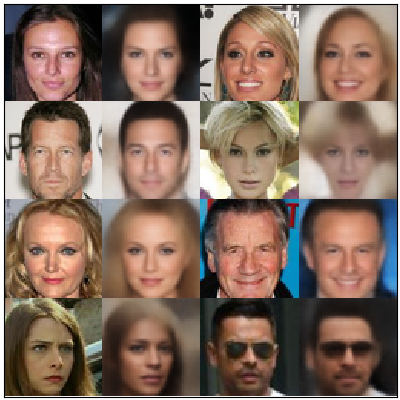} \\  
\end{tabular}
\caption{Results of SCvMAE and SKSAE models trained on CELEB~A dataset. In “test reconstructions” odd rows correspond to the real test points.}
\label{fig:app_celeb_1}
\end{figure}

\begin{figure}[htb]
\centering
\begin{tabular}{@{}c@{}c@{}c@{}c@{}}
 &  Test interpolation &  Test reconstruction &  Random sample \\
 \rotatebox{90}{ \qquad\qquad SWAE} & \,
\includegraphics[height=3cm]{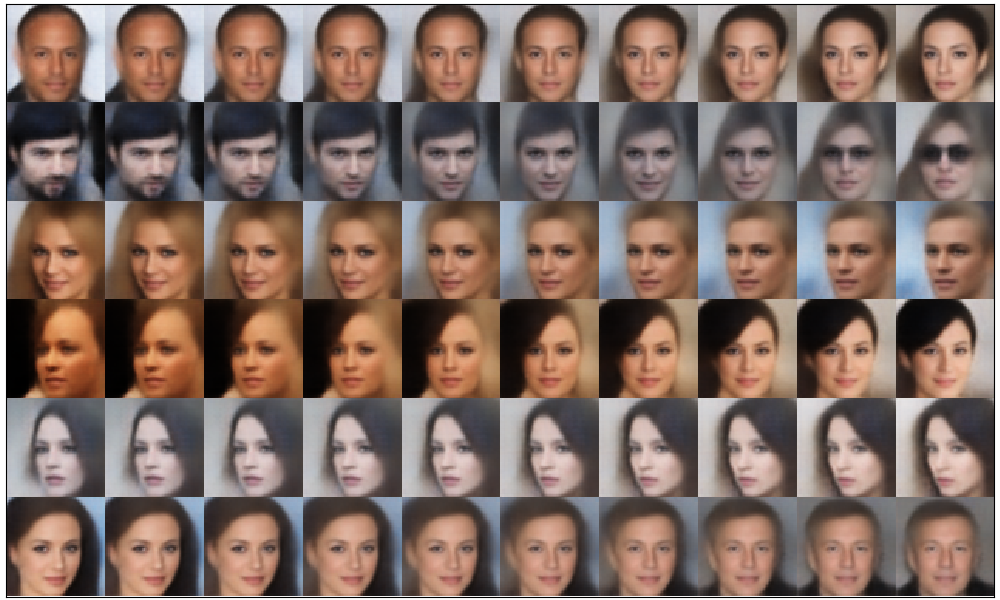} & \,
\includegraphics[height=3cm]{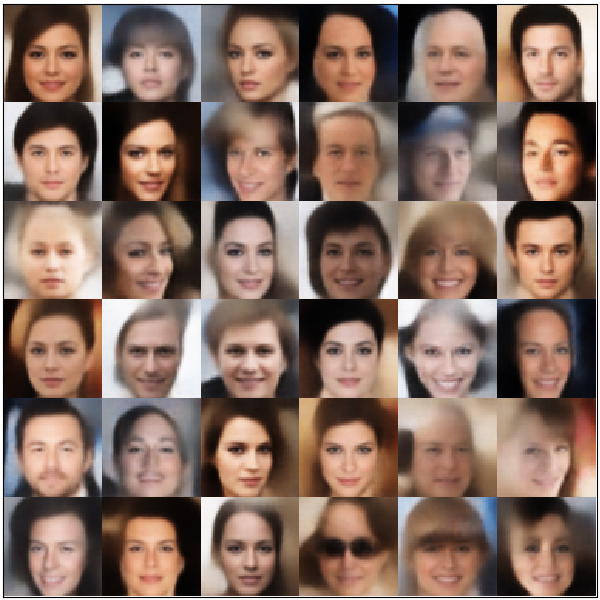}& \, 
\includegraphics[height=3cm]{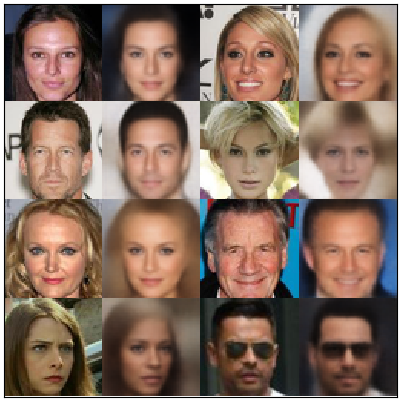} \\ 
 \rotatebox{90}{ \qquad\qquad SCFWAE} & \,
\includegraphics[height=3cm]{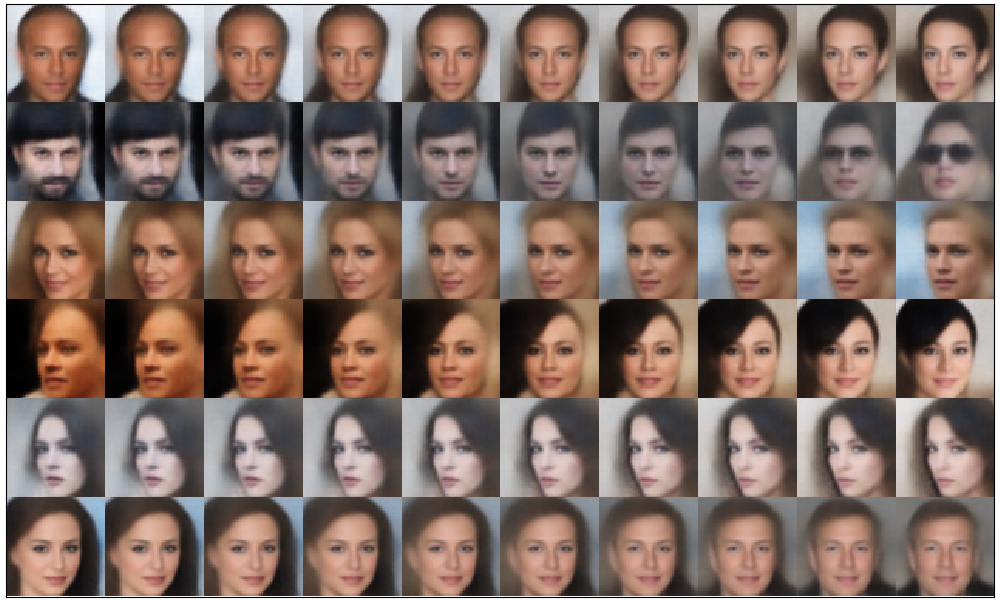} & \,
\includegraphics[height=3cm]{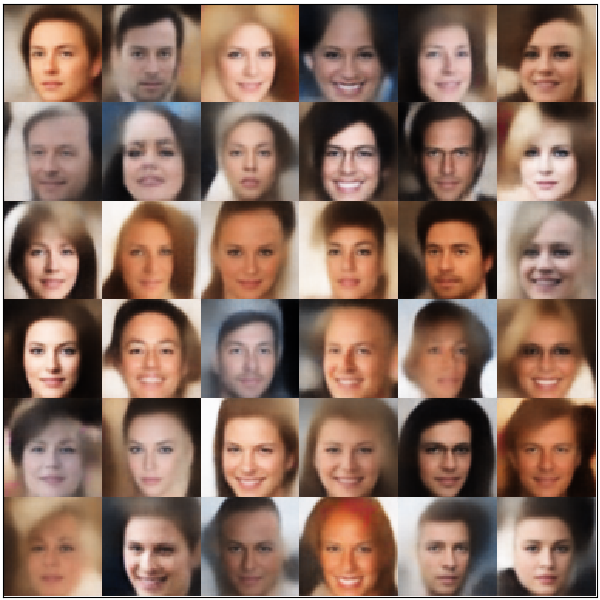}& \, 
\includegraphics[height=3cm]{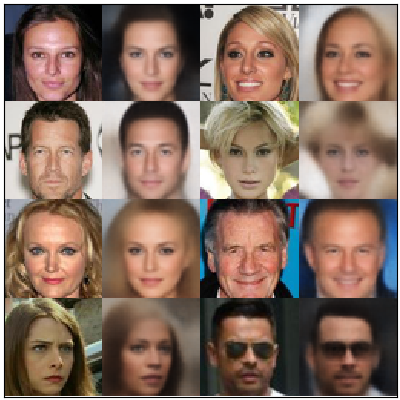} \\ 
 \rotatebox{90}{ \qquad\qquad SCWAE} & \,
\includegraphics[height=3cm]{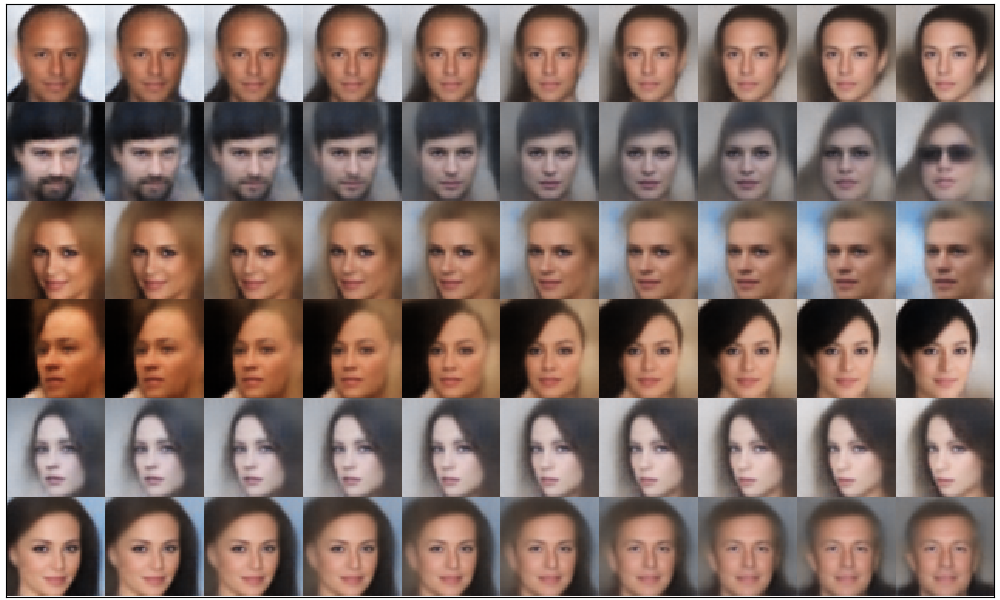} & \,
\includegraphics[height=3cm]{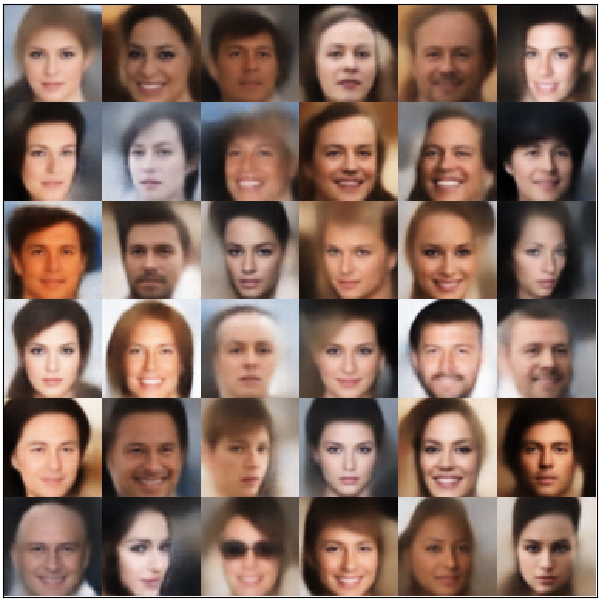}& \, 
\includegraphics[height=3cm]{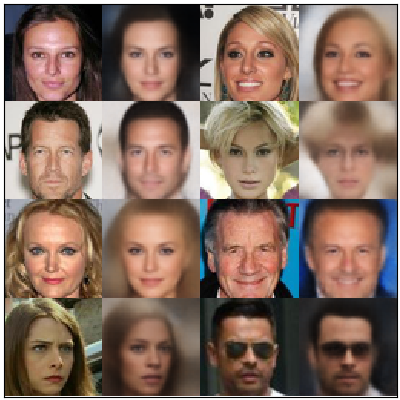} \\ 
\end{tabular}
\caption{Results of SWAE, SCFWAE and SCWAE  models trained on CELEB~A dataset. In “test reconstructions” odd rows correspond to the real test points.}
\label{fig:app_celeb_2}
\end{figure}

Our experiments show that applied methods can be divided into two groups given their generalization properties. The first consists of those which are a modification of standard normality tests:  SCvMAE, SKSAE, see Fig.~\ref{fig:app_celeb_1}, while the second is based on classical distances between samples: SWAE, SCFWAE, SCWAE, see Fig.~\ref{fig:app_celeb_2}. Methods from both groups are correct generative models, but those from the second one give a slightly faster decrease rate of Fr\'{e}chet Inception Distance FID~\cite{heusel2017gans}.

\section{Related works}


The field of representation learning was initially driven by supervised approaches, with impressive results using large labelled datasets. Unsupervised generative modeling, in contrast, used to be a domain governed by probabilistic approaches focusing on low-dimensional data. The situation was changed with introduction of Variational  AutoEncoders (VAE) \cite{kingma2014auto}, which were the first  AutoEncoder based generative models. As a deep learning techniques for learning latent representations, VAE are used to draw images, achieve state-of-the-art results in semi-supervised learning, as well as interpolate between sentences.

One of the the most important aspect in generative models is computational complexity and effectiveness of a distance between the true and the model distribution.
Originally in VAE this computation was carried out using variational methods. An important improvement was brought by using the Wasserstein metric to measure the mentioned distance, which relaxed the need for variational methods and led to the construction Wasserstein  AutoEncoder (WAE)~\cite{tolstikhin2017wasserstein}. 

The next contribution into this research trend was made in \cite{kolouri2018sliced}, where the authors used a sliced version of the Wasserstein distance, instead of the JS-divergency as in WAE-GAN or the maximum mean discrepancy as in WAE-MMD, to penalize dissimilarity between the distribution of encoded training samples and the prior on the latent space. The obtained generative model was called the Sliced-Wasserstein  AutoEncoder (SWAE).

The other related concept can be found in  \cite{tabor2018cramer}, where the authors constructed the Cramer-Wold  AutoEncoder (CWAE), by replacing the sliced Wasserstein distance in SWAE by the newly introduced CW-distance between distributions, which based on the Cramer-Wold Theorem \cite{cramer1936some}. It should be noticed here that, despite the fact that CWAE can be also considered as a version of WAE-MMD method (with a choice of a specific kernel), it involved a closed formula of the CW-distance that came from the application of a sliced approach. Thus, CWAE can be seen as a borderline model between SWAE and WAE-MMD.

With reference to the above mentioned models, in the next section we derive the detailed concept of this paper.

\section{Model}

For convenience of the reader and to establish notation let us start from a classical  AutoEncoder (AE) architecture.
Let $X=(x_i)_{i=1..n} \subset \R^N$ be a given data set, which can be considered as sample from (true but unknown) data distribution $P_X$. The basic aim of AE is to transport the data to a (typically, but not necessarily) less dimensional latent space $\Z=\R^D$ with reconstruction error as small as possible. Thus, we search for an encoder $\E\colon\R^N \to \Z$ and decoder $\D\colon\Z \to \R^N$ functions, which minimize the reconstruction error on the data set $X$:
$$
MSE(X;\E,\D)=\frac{1}{n}\sum_{i=1}^n \|x_i-\D(\E x_i)\|^2.
$$

In turn,  AutoEncoder based generative model is a modification of AE model by introducing a cost function that forces the model to be generative, i.e., ensures that the data transported to the latent space $\Z$ come from the (typically Gaussian) prior distribution $P_\Z$.  A usual way to obtain this is through adding to $MSE(X;\E,\D)$ a regularized (using appropriately chosen hyper-parameter $\lambda>0$) term that penalizes dissimilarity between the distribution of the encoded data $P_{\E(X)}$ and $P_\Z$:
\begin{equation}\label{cost}
COST(X; \E, \D) =  MSE(X;\E, \D) + \lambda\cdot d(P_{\E(X)}, P_\Z).
\end{equation}

The main idea of WAE was based on the use of the Jensen-Shannon divergence (in WAE-GAN) or the maximum mean discrepancy (in WAE-MMD) as $d(P_{\E(X)}, P_\Z)$, which required sampling from $P_\Z$. Note that the Wasserstein metric was applied there to measure only the distance between $P_X$ and the model distribution $P_{\D(\E(X))}$ (this approach is, in fact, a generalization of the reconstruction error $MSE(X;\E,\D)$ and coincide with it in the case of 2nd Wasserstein metric). 

As mentioned in the introduction in this paper, we apply a modification of the cost function, which uses logarithm of the dissimilarity measure instead of (potential grid search over) hyperparameter $\lambda$:
\begin{equation}\label{cost1}
COST(X; \E, \D) =  MSE(X;\E, \D) + \log ( d(P_{\E(X)}, P_\Z)).
\end{equation}

The modification introduced in SWAE relied on the use of the sliced Wasserstein distance to express $d(P_{\E(X)}, P_\Z)$. The main idea was to take the mean of the Wasserstein distances between one-dimensional projections of $P_{\E(X)}$ and $P_\Z$ on a sampled collection of one-dimensional directions. Note that SWAE, similarly to WAE, also needed sampling from $P_\Z$. 
Consequently in SWAE two types of sampling were applied: sampling over one-dimensional projections and sampling from the prior distribution $P_\Z$. The method is effective, but as we show in SCWAE model, it is possible to improve on it by reducing one of the above samplings by using distance between sample and the Gaussian distribution.   

\begin{figure}[htb]
\normalsize
\begin{center}
\includegraphics[width=2.4in]{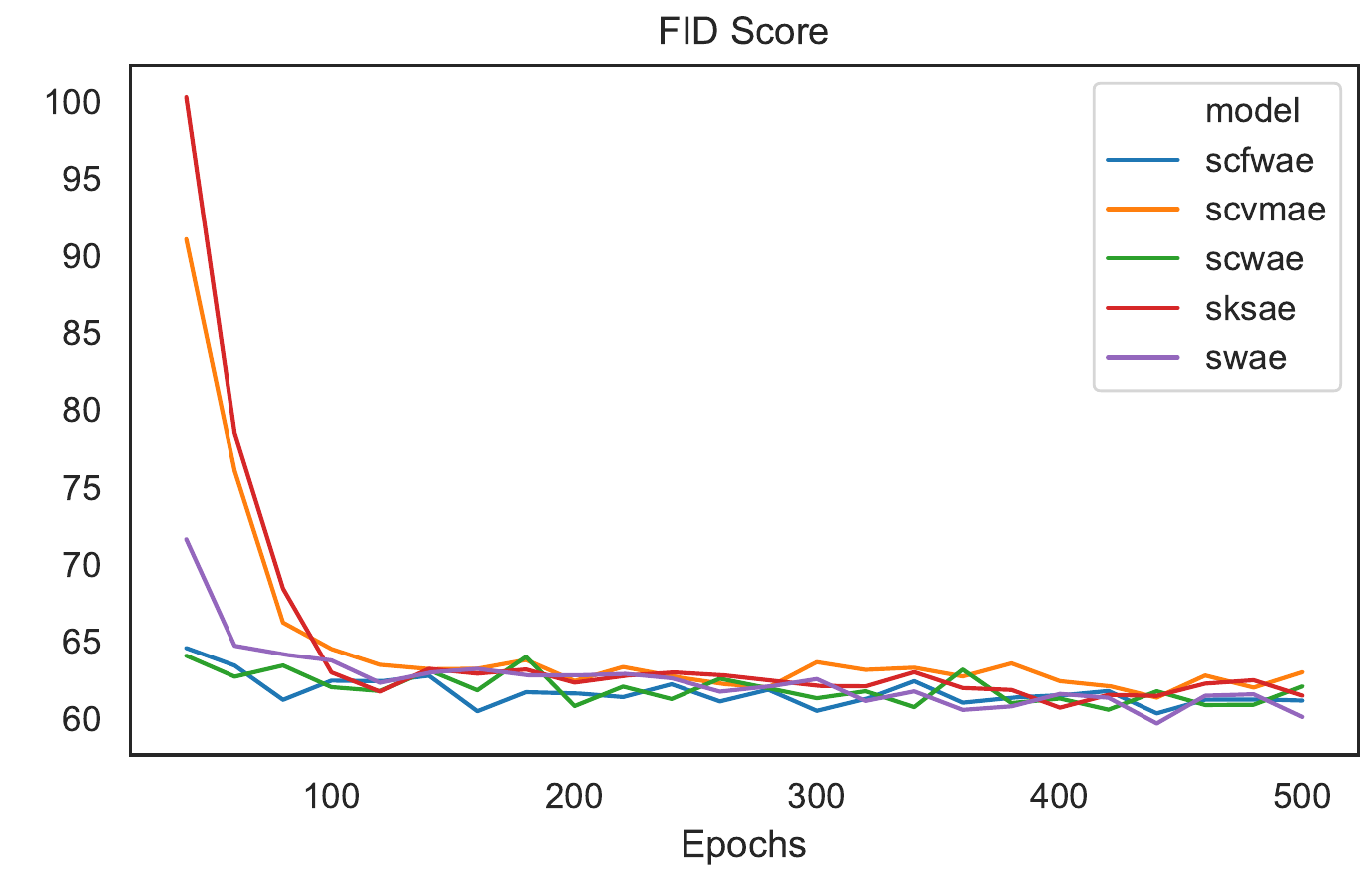} 
\includegraphics[width=2.4in]{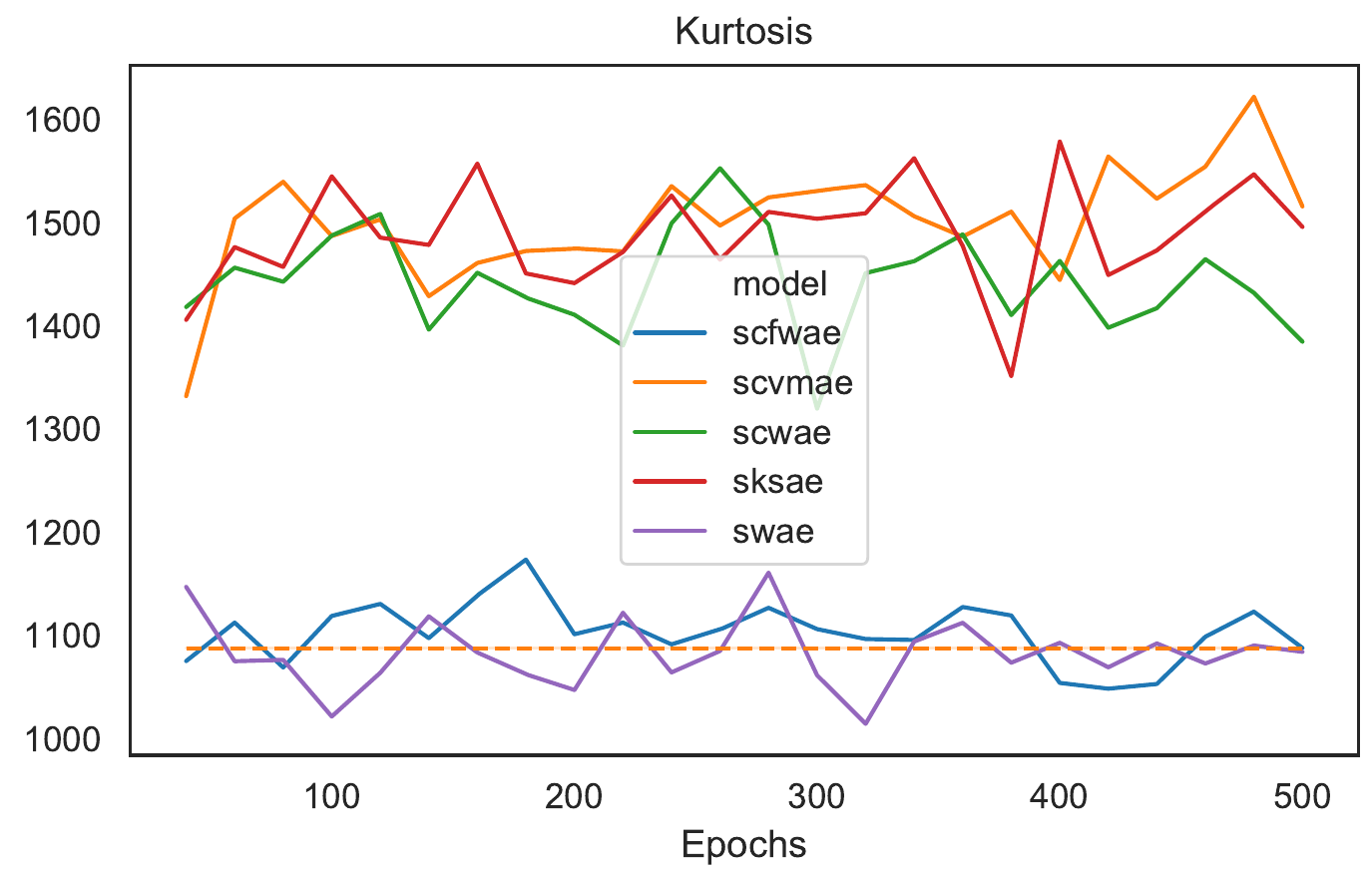}
\includegraphics[width=2.4in]{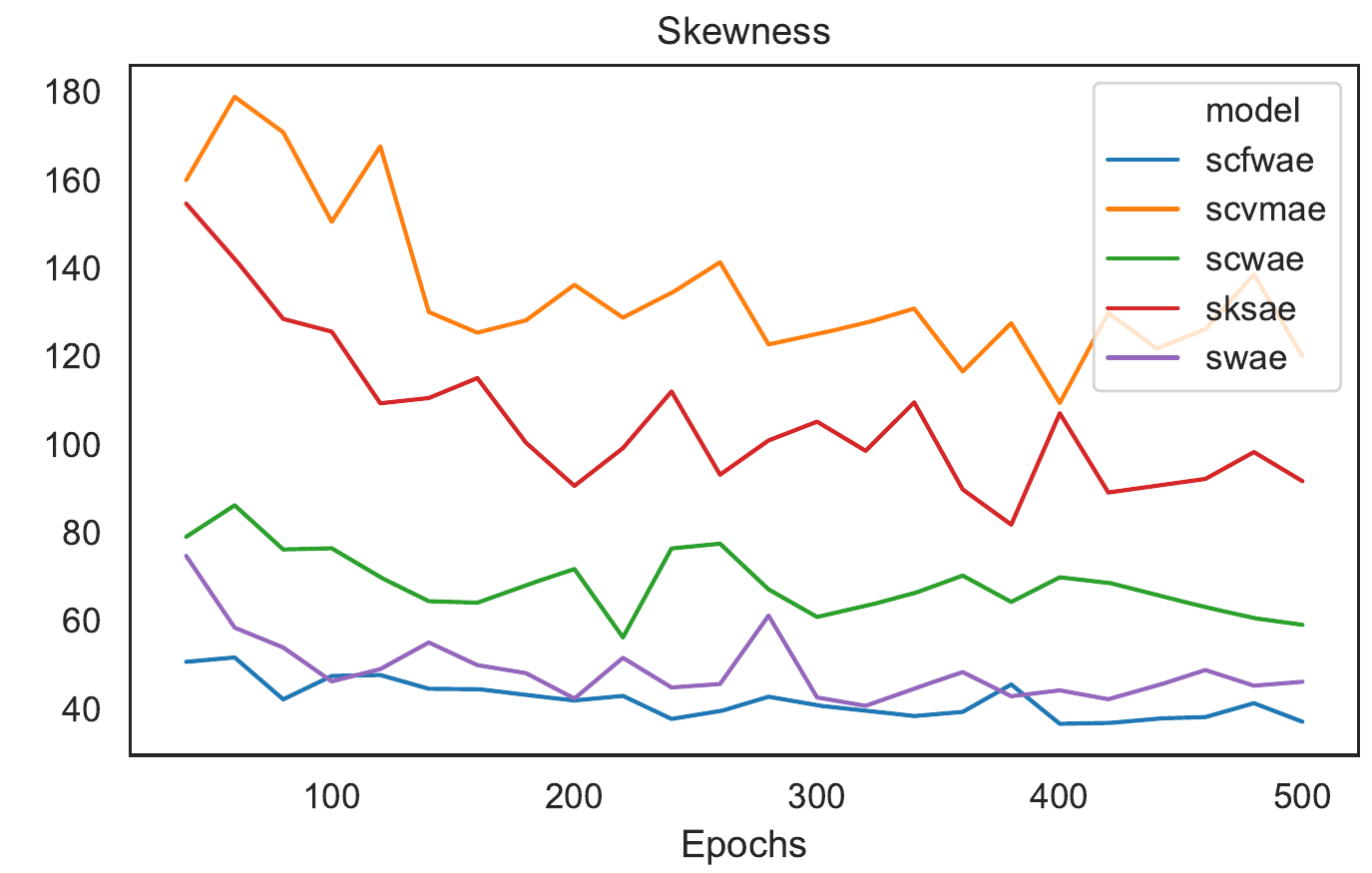}
\includegraphics[width=2.4in]{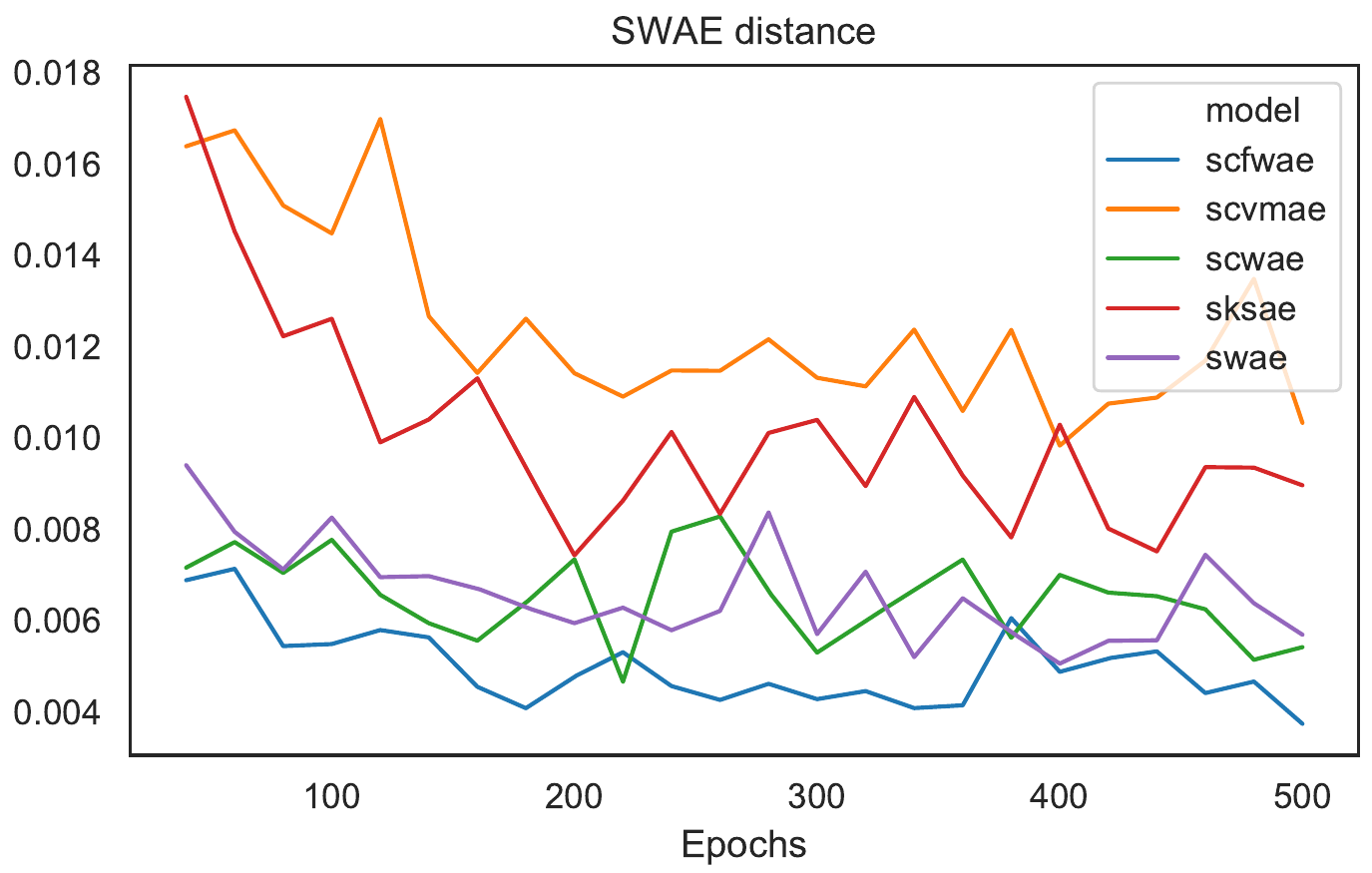}
\end{center}
\caption{Metrics assessing normality of the model output distributions, during training: FID score, Mardia's skewness,  kurtosis and classical SWAE distance of models SCFWAE, SCWAE, SCvMAE, SKSAE and SWAE, on the CELEB~A test set.  Optimal value of kurtosis (i.e. for normal distribution) is given by a dash line.}
\label{fig:conv1}
\end{figure}

To the best of our knowledge, CWAE was the first WAE-like concept that required no sampling.  Assuming the Gaussian  prior $P_\Z$, it used (newly defined) the Cramer-Wold metric to represent $d(P_{\E(X)}, P_\Z)$, which was expressed in an elegant closed form as the distance of a sample from standard multivariate normal distribution $N(0,I)$.

As it was mentioned before, in this paper we examine few variants of sliced distances, which possess computable closed form when considered as a measure of non-normality of a given sample, applied as a penalization term $d(P_{\E(X)}, P_\Z)$, where $P_\Z=N(0,I)$. Specifically, assuming that we have $k$ one-dimensional projections on the spaces spanned by the unit vectors $v_i\in \R^D$ for $i = 1, \ldots, k$, we define:
\begin{equation}\label{dist}
d(P_{\E(X)}, P_\Z) = \frac{1}{k} \sum_{i=1}^k d_{S}( v_i^T  X, N(0,1) ),
\end{equation}
where $d_{S}$ denotes a specified one-dimensional distance function (note that if a random variable $Z\in \R^D$ has the $N(0,I)$ distribution, then $v_i^TZ$ has the $N(0,1)$ distribution). 
%

\section{Dissimilarity measures}

In this section we make few choices of $d_S$'s, which were used (via  \eqref{cost1} and \eqref{dist}) to construct generative  AutoEncoders that are discussed in this paper.




\textbf{Sliced Wasserstein  AutoEncoder (SWAE).} 
In the original SWAE paper \cite{kolouri2018sliced}, to express $d_S$ the authors use the square of the 2-nd Wasserstein distance between the (empirical) distributions generated by the respective samples. 
This leads to the following formula:
\begin{align*}
d_{S}( Y, Z)&=\int_0^1 (P^{-1}_{Y}(t)- P^{-1}_{Z}(t))^2\, dt =
\int_{0}^1 \left(\sum_{i=1}^n (y_{(i)}-z_{(i)})\mathbf{1}_{\frac{i-1}{n}<t\leq \frac{i}{n}}\right)^2\,dt\nonumber\\
&=\frac{1}{n}\sum_{i=1}^n (y_{(i)}-z_{(i)})^2,
\end{align*}
where $P^{-1}_{*}(t)=\inf\{x\in \R: P_{*}(x)\geq t\}$ for $t\in (0,1)$, whereas $(y_{(1)},\ldots, y_{(n)})$ is an ordered 
sample $Y=(y_1,\ldots, y_n)$ and $(z_{(1)},\ldots, z_{(n)})$ represents an ordered sample $Z=(z_1,\ldots, z_n)$ derived from $N(0,1)$.

\textbf{Sliced Closed Form Wasserstein  AutoEncoder (SCFWAE).} 
In the original SWAE paper authors have used Wasserstein distance between samples~\cite{kolouri2018sliced}. We show in SCFWAE a model that we can simplify it by  using distance between sample and Gaussian density distribution (consequently, no sampling from the normal distribution is necessary).
We define $d_S$ as the square of the 2nd Wasserstein distance:
\begin{align*}
d_{S}( Y, N(0,1))&=\int_0^1 (P^{-1}_{Y}(t)- P^{-1}_{0}(t))^2\, dt =
\int_{0}^1 \left(\sum_{i=1}^n y_{(i)}\mathbf{1}_{\frac{i-1}{n}<t\leq \frac{i}{n}}-P^{-1}_{0}(t)\right)^2\,dt\nonumber\\
&=\frac{1}{n}\sum_{i=1}^n y_{(i)}^2
- 2\sum_{i=1}^n y_{(i)}\int_{\frac{i-1}{n}}^{\frac{i}{n}} P^{-1}_{0}(t)\,dt + \int_{-\infty}^\infty y^2\cdot p_0(y)\,dy\nonumber\\
&= \frac{1}{n}\sum_{i=1}^n y_{(i)}^2 - 2\sum_{i=1}^n y_{(i)}\int_{Q_\frac{i-1}{n}}^{Q_\frac{i}{n}} y\cdot p_0(y)\,dy + 1\\
&= 1+\frac{1}{n}\sum_{i=1}^n y_{(i)}^2 - \sqrt{\frac{2}{\pi}}\sum_{i=1}^n y_{(i)}\int_{Q_{\frac{i-1}{n}}}^{Q_{\frac{i}{n}}}y\cdot \exp(-\frac{y^2}{2})\,dy\\
&= 1+\frac{1}{n}\sum_{i=1}^n y_{(i)}^2 + \sqrt{\frac{2}{\pi}}\sum_{i=1}^n y_{(i)}(\exp(-\frac{1}{2}Q_{\frac{i}{n}}^2)-\exp(-\frac{1}{2}Q_{\frac{i-1}{n}}^2)),
\end{align*}
where 
$P_0$, $p_0$, and $Q_r$ denote the distribution function, the density function and the $r$-th quantile of $N(0,1)$.

\textbf{Sliced Cramer-Wold  AutoEncoder (SCWAE).}
Following \cite{tabor2018cramer}, as $d_S$ we choose the square of the one dimensional Cramer-Wold distance, which is defined as an $\ell_2$ distance between a sample $Y=(y_1,\ldots, y_n)\subset \R$ and $N(0,1)$, both smoothen using a Gaussian kernel $N(0,\gamma)$, where $\gamma=(\tfrac{4}{3n})^{2/5}$ is a bandwidth constant given by the Silverman's rule of thumb (see~\cite{silverman1986density}). This leads to the following formula:
\begin{align*}
    d_{S}(Y, N(0,1))&=
    \big\|\frac{1}{n}\sum \limits_{i=1}^n p_{y_i,\gamma} -  p_{0,1+\gamma} \big\|^2_2  
    =\frac{1}{n^2}\il{\sum \limits_{i=1}^n p_{y_i,\gamma},\sum \limits_{i=1}^n p_{y_{i},\gamma}}_2 \\ 
    & + \il{ p_{0,1+\gamma},p_{0,1+\gamma}}_2    {-
     \frac{2}{n}\il{\sum \limits_{i=1}^n p_{y_i,\gamma},p_{0,1+\gamma}}_2} \\ 
    & =\frac{1}{n^2}\sum \limits_{i,j=1}^n p_{y_i-y_{j},2\gamma}(0)+ p_{0,2+2\gamma}(0)-\frac{2}{n}\sum \limits_{i=1}^n p_{y_i,1+2\gamma}(0),
\end{align*}
where by $p_{m,\sigma}$ we denote the density function of $N(m,\sigma)$.


In addition to the classic distances used in generative models, we can use various dissimilarity measures related to classical statistical tests. In the literature there are many tests for normality, which work well in the case of one dimensional datasets. In the paper we verify a possibility of application of that classical statistical models in deep generative architectures.

\textbf{Sliced Cram\'er-von Mises  AutoEncoder (SCvMAE).}
The first statistical model we apply is the Cram\'er-von Mises test for normality. It can be easily derived from an application of the Wasserstein distance. Indeed, 
basing on the known fact that if $Y$ is a random variable then the variable $P_Y(Y)$ has the continuous uniform distribution $U(0,1)$, as $d_S$ we use the square of the 2nd Wasserstein distance between the distribution of $P_Y(Y)$ and $U(0,1)$, i.e.:
\begin{align}\label{cm}
d_S(Y,N(0,1)) & =  
\int_0^1 (P^{-1}_{Z}(t)- P^{-1}_{1}(t))^2\, dt \nonumber \\
& =  \int_{0}^1 \left(\sum_{i=1}^n z_{(i)}\mathbf{1}_{\frac{i-1}{n}<t\leq \frac{i}{n}}-P^{-1}_{1}(t)\right)^2\,dt\nonumber \\
& =  \frac{1}{n}\sum_{i=1}^n z_{(i)}^2 - 2\sum_{i=1}^n z_{(i)}\int_{\frac{i-1}{n}}^{\frac{i}{n}}t\,dt +\frac{1}{4}+\frac{1}{12} \\
& =  \frac{1}{n}\sum_{i=1}^n z_{(i)}^2- \frac{1}{n^2}\sum_{i=1}^n z_{(i)}\nonumber 
\cdot (i^2-(i-1)^2)+\frac{1}{3} \\ \nonumber & =  \frac{1}{n}\sum_{i=1}^n z_{(i)}^2+\frac{1}{n^2}\sum_{i=1}^n z_{(i)}(2i-1)+\frac{1}{3},
\end{align}
where $P_1$ is the distribution function of $U(0,1)$ and $(z_{(1)},\ldots,z_{(n)})$ is an ordered sample $Z=(P_{1}(y_1),\ldots, P_{1}(y_n))$. Then it is easy to verify (see, e.g., \cite{mazur2018on}) that \eqref{cm} coincides with the Cram\'er-von Mises distance between $P_{Y}$ and $P_{0}$, which is used in the Cram\'er-von Mises goodness of fit test for normality.

\textbf{Sliced Kolmogorov-Smirnov  AutoEncoder (SKSAE).}
Our last choice of $d_S$ is a clasical Kolmogorov-Smirnov distance, which is used as a statistics in the Kolmogorov-Smirnov goodness of fit test for normality. It is expressed (see, e.g.,~\cite{hazewinkel2001kolmogorov}) by the following formula:
$$
d_S(Y,N(0,1)) =\sup_{y} \big|P_Y(y)-P_0(y)\big|=\max_{i} \big|\tfrac{i}{n}-P_0(y_{(i)})\big|.
$$

\begin{figure}[htb]
\normalsize
\begin{center}
\includegraphics[width=2.4in]{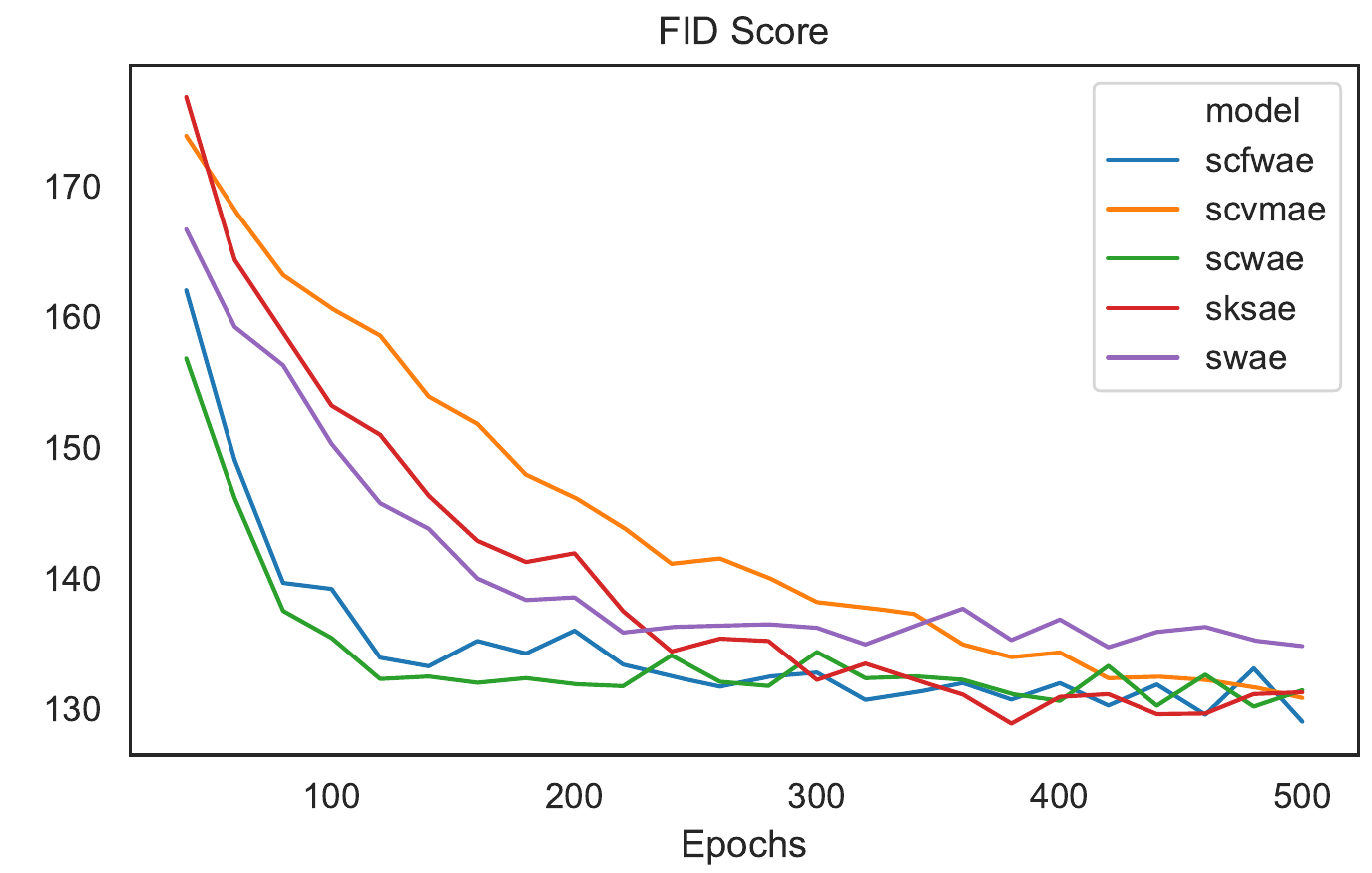} 
\includegraphics[width=2.4in]{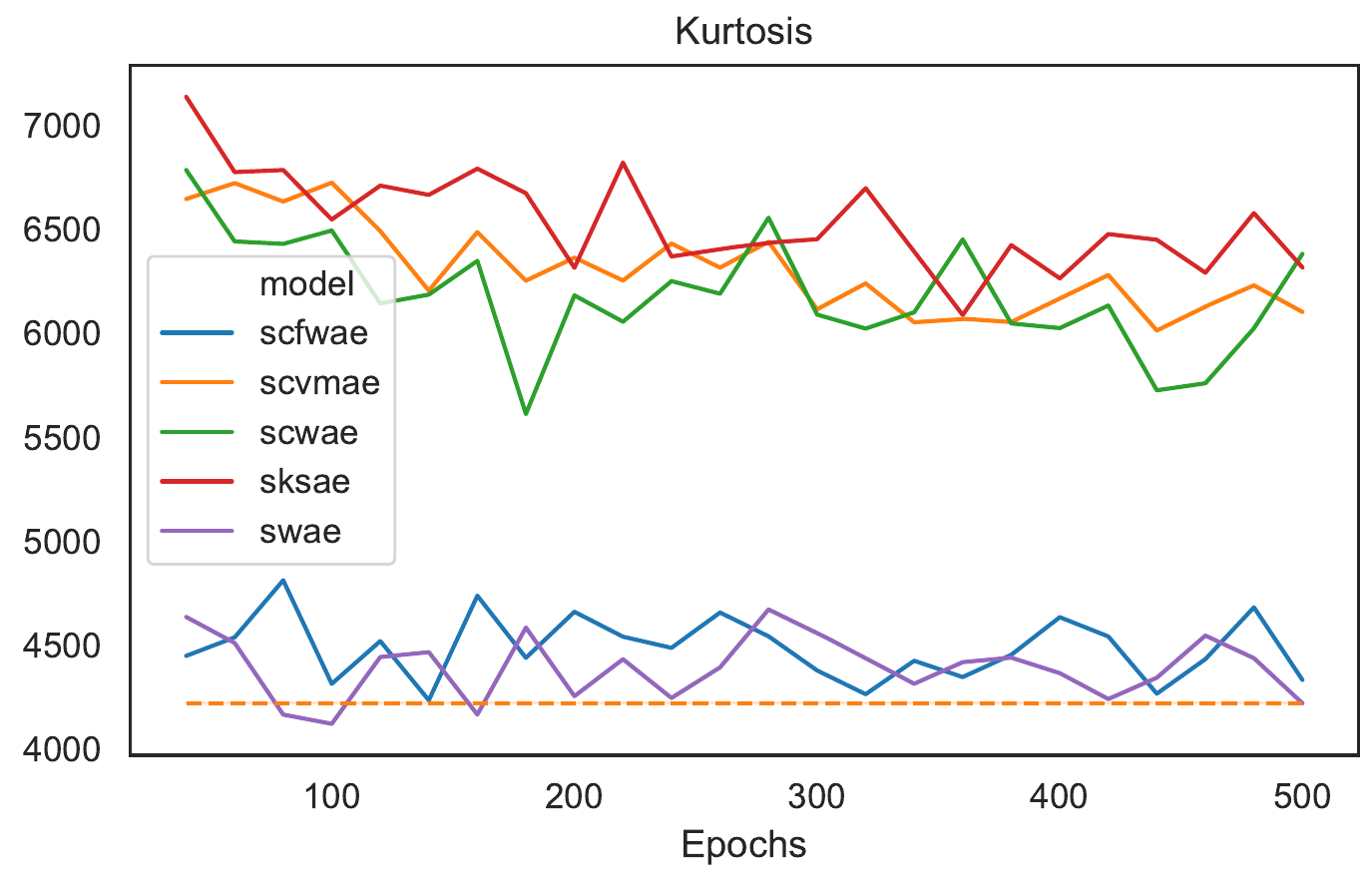}
\includegraphics[width=2.4in]{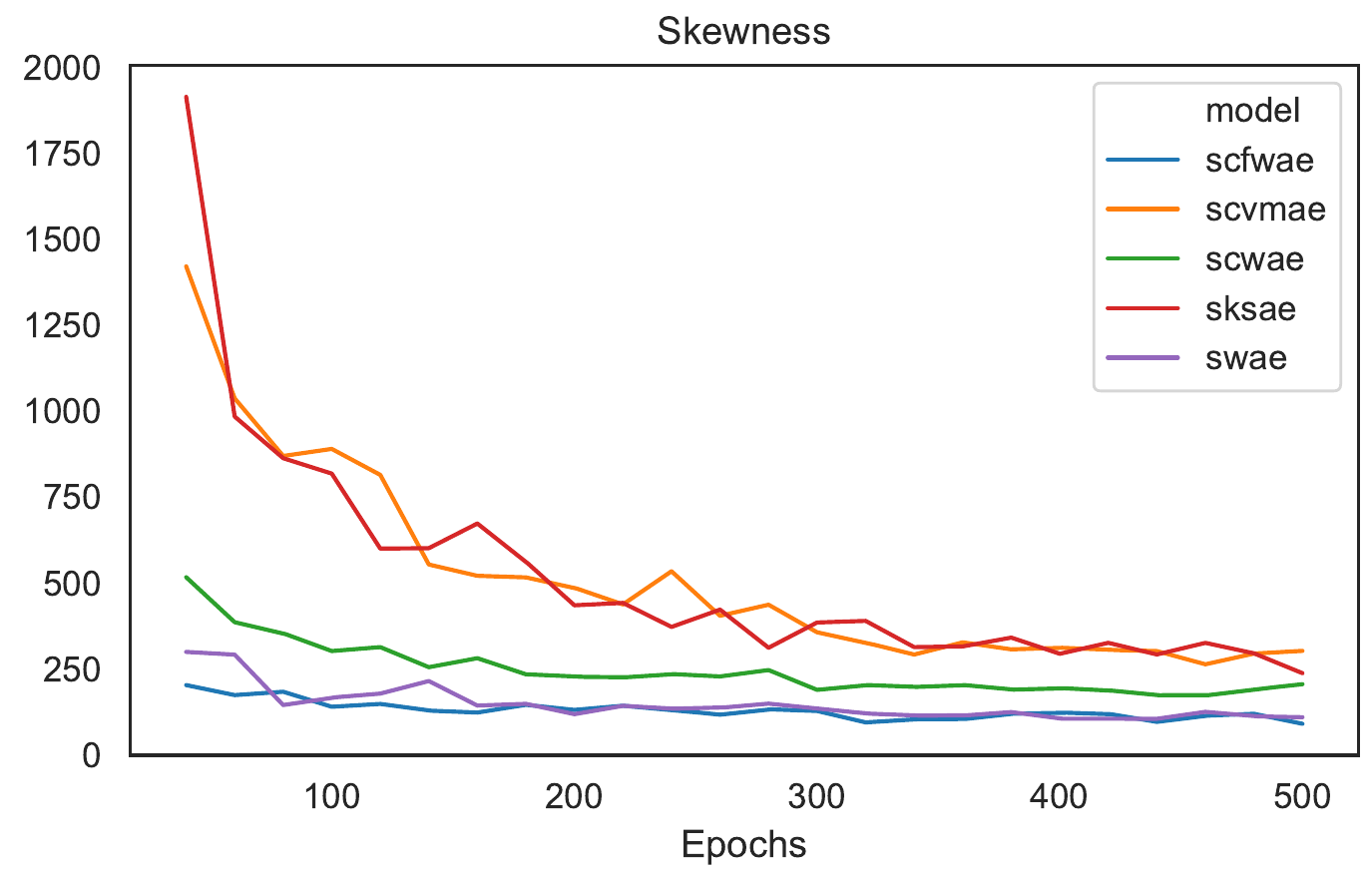}
\includegraphics[width=2.4in]{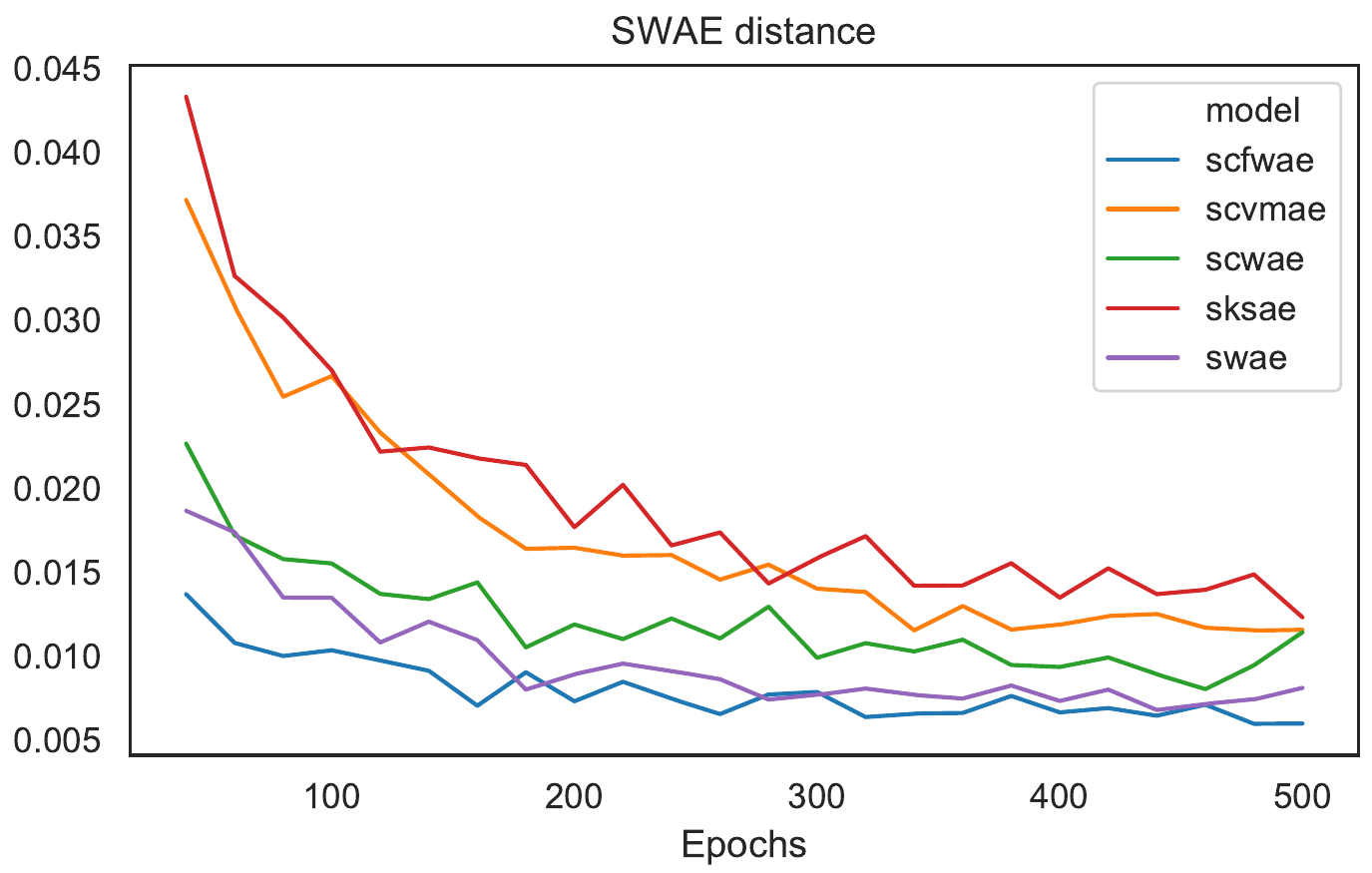}
\end{center}
\caption{Metrics assessing normality of the model output distributions, during training: FID score, Mardia's skewness,  kurtosis and classical SWAE distance of models SCFWAE, SCWAE, SCvMAE, SKSAE and SWAE, on the Cifar~10 test set.  Optimal value of kurtosis, (i.e. for normal distribution) is given by a dash line.}
\label{fig:conv2}
\end{figure}

\section{Experiments}

In this section we shall empirically validate proposed models on standard benchmarks for generative models CELEB~A, CIFAR-10 and MNIST. We will compare different approaches to sliced generative models SCFWAE, SCWAE, SCvMAE, SKSAE, SWAE with classical  SWAE~\cite{kolouri2018sliced}. 
As we shall see, all the above methods can be divided in to two groups. The first contains methods which are all modifications of classical normality tests: SCvMAE, SKSAE, while the second one those based on classical distances between multidimensional samples: SCFWAE, SCWAE and classical SWAE. It can be noticed that the second class of methods gives a slightly better results.

In the experiments we use two basic architecture types. Experiments on MNIST use a feed-forward network for both encoder and decoder, and a 20 neuron latent layer, all using ReLU activations. For CIFAR-10 and CELEB~A data sets we use convolution-deconvolution architectures. 

The quality of a generative model is typically evaluated by examining generated samples or by interpolating between samples in the latent space. We present such a~comparison between all approaches in Fig.~\ref{fig:app_celeb_1} and Fig.~\ref{fig:app_celeb_2}. The experiment shows that there are no perceptual differences between considered models. In order to quantitatively compare all above slicing methods we use three measures.
First of all, we use the Fr\'{e}chet Inception Distance (FID)~\cite{heusel2017gans}, which is the most popular measure of generalization in deep generative models.

Next, following experiments from \cite{tabor2018cramer}, we verified standard normal distribution in the latent by using statistical normality tests, i.e. Mardia tests~\cite{henze2002invariant}. More precisely we use skewness $b_{1,D}(\cdot)$ and kurtosis $b_{2,D}(\cdot)$ of a sample $X=(x_i)_{i=1..n} \subset \R^D$:
$$
\begin{array}{c}
b_{1,D}(X)=\tfrac{1}{n^2}
\sum \limits_{j,k}(x_j^Tx_k)^3 \text{ and }
b_{2,D}(X)=\tfrac{1}{n}
\sum \limits_{j}\|x_j\|^4
\end{array}
$$
are close to that of standard normal density. The expected Mardia’s skewness and kurtosis for standard multivariate normal distribution is $0$ and $D(D + 2)$, respectively.


\begin{table}[htb]
\small
\caption{Comparison between different models output distributions and the normal distribution, together with reconstruction error. All model outputs except AE are similarly close to the normal distribution. Normality is assessed by comparing Mardia's skewness, kurtosis (normalized), and the reconstruction error. For reference FID scores are provided as well (except for MNIST, where it is not defined). }
\begin{center}
\begin{tabular}[width=\textwidth]{llrrrrr}  
\toprule
\hspace{-1ex} Data set \hspace{-1ex} & Method & SWAE & SKSAE & SCWAE & \hspace{-1ex}SCvMAE & \hspace{-1ex}SCFWAE\hspace{-1ex} \\
\midrule             
MNIST    & Skewness      & 35.86 & 57.34     & 34.19 &  59.22  & 37.41 \\
& Kurtosis (normalized)  & -57.46 & 35.33  & -10.29 &  23.82 & -31.93 \\
      & Reconstruction 
                 error   & 5.37  & 5.01      & 5.35  &  5.04  &  5.42  \\
\midrule             
CIFAR10    & Skewness      & 110.49 & 238.52     & 206.50 &  303.45  & 91.42 \\
& Kurtosis (normalized)  & -0.96 & 2093.68  & 2159.31 &  1879.98 & 111.21 \\
      & Reconstruction 
                 error   & 27.02  & 24.93      & 27.29  &  25.60  &  26.35  \\ 
& \em{FID score
                 error}   & 134.87       & 131.32 & 131.48  &  130.89 & 129.07 \\                 
\midrule              
CelebA    & Skewness      & 46.14 & 91.68     & 59.07 &  120.09  & 37.07 \\
& Kurtosis (normalized)  & -3.60 & 408.24  & 296.99 &  428.18 & 0.17 \\
      & Reconstruction 
                 error   & 115.68  & 115.57      & 115.30  &  115.62  &  115.26  \\ 
& \em{FID score
                 error}   & 60.10       & 61.49 & 62.09  &  63.01 & 61.16 \\                 
\bottomrule
\end{tabular}
\end{center}
\label{tab:comp}
\end{table}

Results are presented in Figure~\ref{fig:conv1}, Figure~\ref{fig:conv2} and Table~\ref{tab:comp}. In Figure~\ref{fig:conv1} we report for CELEB~A data set the value of FID score, Mardia's skewness and kurtosis during learning process of SCFWAE, SCWAE, SCvMAE, SKSAE, SWAE (measured on the validation data set). Methods based on  modification of classical normality tests: SCvMAE, SKSAE obtain a sightly worse skewness and kurtosis in the case of both data-sets. On the other hand all methods gives similar level of FID score but it can be seen that SCFWAE, SCWAE and classical SWAE faster convergence.

\section{Conclusions}

In this paper, we have compared a few different approaches to construct sliced AutoEncoder based generative models. In particular, we used classical one-dimensional distances between samples and arbitrary fixed density distribution, some of them derived from classical (one-dimensional) goodness of fit tests for normality. Moreover, we have constructed SCFWAE -- a simplified version of SWAE, where there is no necessity to sample from the normal prior. Our experiments show that all considered method are correct generative models, but the methods based on the Wasserstein and the Cramer-Wold  distances have slightly faster decrease rate of the FID score.



\end{document}